\documentclass[letterpaper, 10 pt, conference]{ieeeconf}  

\IEEEoverridecommandlockouts

\usepackage{graphicx}
\usepackage{xcolor}
\usepackage{subcaption}
\usepackage{float}
\usepackage{algorithm}
\usepackage{algorithmic}
\usepackage{amsmath}
\usepackage{siunitx}
\PassOptionsToPackage{hyphens}{url}\usepackage{hyperref} 
\usepackage{comment}
\usepackage{colortbl}
\usepackage{gensymb}
\usepackage{xcolor}
\usepackage{balance}

\title{\LARGE \bf
Terrain-Aware Adaptation for Two-Dimensional UAV Path Planners
}

\author{Kostas Karakontis$^{1}$, Thanos Petsanis$^{1,2}$, Athanasios Ch. Kapoutsis$^{2}$, \\Pavlos Ch. Kapoutsis$^{2}$, Elias B. Kosmatopoulos$^{1,2}$
\thanks{$^{1}$Department of Electrical and Computer Engineering,
Democritus University of Thrace, Xanthi, 67100, Greece
        {\tt\small konskara41@ee.duth.gr, apetsani@ee.duth.gr, kosmatop@ee.duth.gr}}%
\thanks{$^{2}$Information Technologies Institute, The Centre for Research \& Technology, Thessaloniki, 57001, Greece
        {\tt\small athpets@iti.gr, athakapo@iti.gr, pkapoutsis@mail.iti.gr, kosmatop@iti.gr }}%
}

\begin{document}

\maketitle
\thispagestyle{empty}
\pagestyle{empty}

\begin{abstract}
Multi‑UAV Coverage Path Planning (mCPP) algorithms in popular commercial software typically treat a Region of Interest (RoI) only as a 2D plane, ignoring important 3D structure characteristics. This leads to incomplete 3D reconstructions, especially around occluded or vertical surfaces. In this paper, we propose a modular algorithm that can extend commercial two-dimensional path planners to facilitate terrain-aware planning by adjusting altitude and camera orientations. To demonstrate it, we extend the well‑known DARP (Divide Areas for Optimal Multi‑Robot Coverage Path Planning) algorithm and produce DARP-3D. We present simulation results in multiple 3D environments and a real‑world flight test using DJI hardware. Compared to baseline, our approach consistently captures improved 3D reconstructions, particularly in areas with significant vertical features. An open-source implementation of the algorithm is available here: \href{https://github.com/konskara/TerraPlan}{https://github.com/konskara/TerraPlan}
\end{abstract}

\section{INTRODUCTION}
\subsection{Motivation}

In recent years, UAVs have been employed for a wide range of applications such as search \& rescue, reconnaissance \& surveillance, and more importantly, surveying \& mapping. Their growing popularity can be attributed to ease of operation, reduced costs, and increased flexibility \cite{drones_on_rise}. What is more, persistent demand for UAVs fuels rapid technological advancements, continuously enhancing their capabilities, ease of use, and affordability \cite{review_UAVs}.

Particularly, multi-UAV Coverage Path Planning (mCPP), which aims at maximizing area coverage from multiple UAVs, has become more dominant in literature, but has yet to be supported by many UAV or service providers, such as DJI, Pix4D, DroneDeploy, among others. Their embedded, undisclosed algorithms seem unintelligent as they produce standard grid-based paths. Furthermore, these algorithms as well as many research works \cite{Overcome_FOMO, med1, dynamic_2D_DRL}, treat a Region of Interest (RoI) only as a 2D plane by outputting paths with fixed altitudes and do not account for diverse height profiles. Such assumptions are problematic for applications that require comprehensive 3D data collection, like area inspection or detailed 3D model reconstructions \cite{crack_detection, med2, bridge_inspection}. Since the resulting flight plans have a fixed altitude and the camera is directed to the ground below, UAVs can only observe the top surfaces of the terrain, leaving occluded or vertical structures unobserved or insufficiently captured. Among pilots it is a standard practice to take manual control of UAVs to capture images from different angles or change the gimbal pitch to an inclined angle and even employ double-grid paths (which take double the time) in an attempt to capture more angles of the surface.
 
The approach proposed in this paper represents a first step toward bridging that gap by enabling effective 3D mapping that can be seamlessly integrated into existing commercial tools. Specifically, we build upon the well-known, state-of-the-art 2D mapping algorithm called the Divide Areas Algorithm for Optimal Multi-Robot Cover Path Planning (DARP)\cite{DARP}, perform evaluation on 4 synthetic testbeds and then, using DJI's API, execute a  real-world mission. 

\begin{figure}[t!]
    \centering
    \begin{subfigure}[t]{0.49\linewidth} 
        \centering
        \vspace{0pt} 
       \includegraphics[width=1\linewidth, height=1\linewidth]{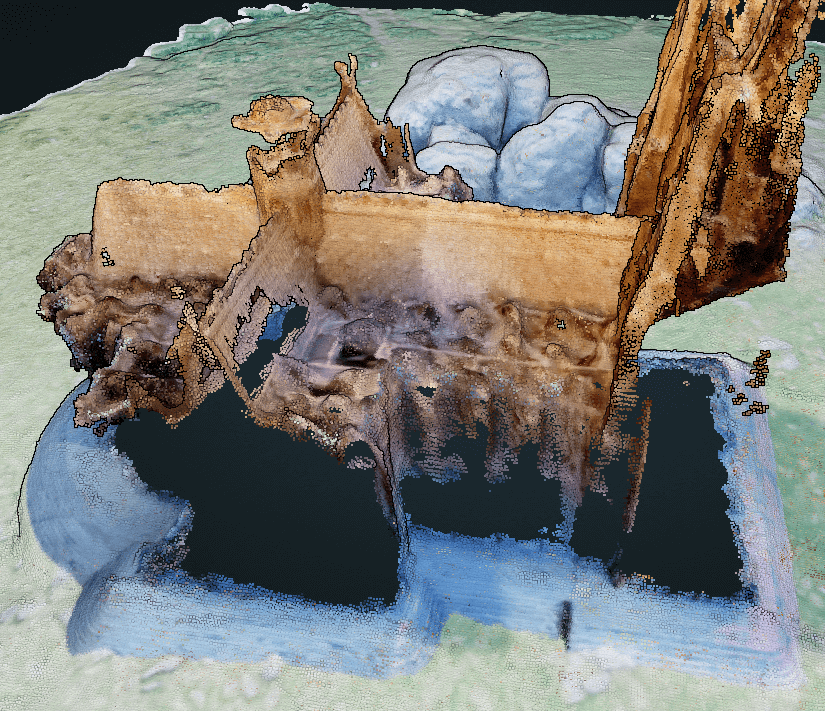} 
        \caption{DARP model}
        \label{fig:DARP_viewpoints}
    \end{subfigure}
    \hfill
    \begin{subfigure}[t]{0.49\linewidth} 
        \centering
        \vspace{0pt} 
         \includegraphics[width=1\linewidth, height=1\linewidth]{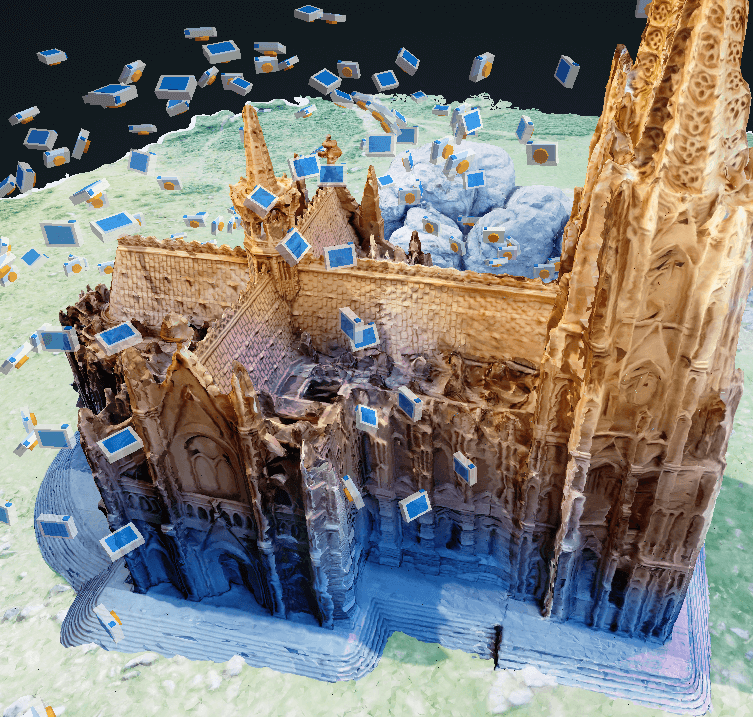} 
        \caption{DARP-3D model}
        \label{fig:DARP3D_viewpoints}
    \end{subfigure}
    \caption{Model side-by-side comparison of the Cathedral testbed. Baseline algorithm DARP model (subfigure a) and the same algorithm equipped with our approach, named DARP-3D (subfigure b). Position and orientation of viewpoints (the blue rectangles) are adjusted to improve 3D reconstruction of the object of interest.}
    \label{fig:viewpoints}
\end{figure}

\subsection{Related Work}
Existing methods in 3D mapping and coverage path planning often focus on single-UAV or object-centric strategies. For instance, PredRecon \cite{Predrecon} is a state-of-the-art algorithm designed for high-fidelity 3D reconstructions of individual objects, but it does not readily scale to larger areas and supports only a single UAV. Likewise, FC-Planner \cite{FC_planner} excels at inspecting complex structures but remains limited to single-UAV operations and narrowly defined environments.  

On the other hand, mCPP is a much more complex problem to solve, since it requires reliable task distribution and even swarm awareness, but enables more applications, better efficiency \cite{review_UAVs} and has shown better performance for 3D reconstruction \cite{review_3D_UAVs}. The same researchers of PredRecon and FC-Planner extended their algorithms to utilize multiple UAVs in SOAR\cite{SOAR},  More and more works are emerging which try to tackle the mCPP problem with different approaches. Due to its complexity, these works often involve heuristic methods such as evolutionary algorithms \cite{evolutionary}, ant-colony optimization \cite{ant_colony}, or simpler ones like A-star\cite{Huang_DARP}. DARP \cite{DARP} is one such algorithm. Having been used in various domains \cite{Zhang, BINN, Chang_DARP}, even in the real-world \cite{ship_DARP}, with extensions\cite{Huang_DARP, Alessandro_DARP} or as comparison \cite{Chang_DARP, Alessandro_DARP, tmstc, greedy_entropy, turn_reduction}, it is a well documented and supported algorithm that has shown low processing time and high coverage percentage performance. Certain aspects of it are already included in both commercial and open-source platforms \cite{DARP_app}.

Related work can also be categorized as offline and online algorithms. Offline means that the paths are not adapted mid-flight, but instead are calculated beforehand usually given known information about the environment. Many approaches conduct a quick first-stage mission which is intended to capture significantly less data from a region. Then they construct an abstract environment representation, serving as the bases model, upon which optimal paths can be computed. In \cite{semantic_3D_UAVs} researchers employ semantic segmentation to identify contextual information. In this way, they can exclude segments of no interest such as cars, roads, forests etc. and instead compute waypoints around buildings. They do so, by designing a reward equation and pick out the viewpoints that maximize it in a Monte-Carlo fashion. Researchers in \cite{Wang} propose a cooperative multi-UAV flight planning method to efficiently reconstruct high-resolution 3D building models. By precomputing flight paths based on building geometry, UAV endurance, and imaging constraints, they minimize redundant image capture while ensuring full coverage.

While numerous others advanced methods have been proposed for comprehensive 3D scanning of surfaces, none have achieved widespread adoption in mainstream commercial software. Obstacles include algorithmic complexity, disclosed code, and the need for on-board processing, which impose significant hardware and operational requirements. Many rely on fixed altitudes or a single camera orientation, making it difficult to capture occlusions or vertical structures. As a consequence, even algorithms that produce impressive results face hurdles which limit large-scale, practical deployment.

\subsection{Contributions}
In this work, we provide a practical solution for unifying high-quality 3D capture with the reliability and simplicity that current multi-UAV missions demand. We showcase it by extending DARP into DARP-3D with adjusted altitudes and intelligent camera orientations. Specifically, our contributions are:
\begin{itemize}
    \item A modular extension for any fixed-altitude path planner that seeks to improve its 3D reconstruction capabilities
    \item An efficient mCPP algorithm that also has high 3D reconstruction performance and
    \item An extensive simulation study and real-world validation on a commercial UAV platform, demonstrating the practicality and effectiveness of the proposed system.
\end{itemize}

In the following section, Section \ref{sec:methodology} we elaborate on the steps of our pipeline and on Section \ref{sec:results} we: a) describe the simulation setup of four challenging synthetic testbed environments b) elaborate on the evaluation process, c) provide metrics compared to our baseline algorithm and d) perform a real-world flight test. Finally, in Section \ref{sec:conclusion} we discuss main outcomes, further possible applications and future work.

\section{Methodology}
\label{sec:methodology}

\begin{figure*}[t!]
    \centering    
    \includegraphics[width=0.7\textwidth]{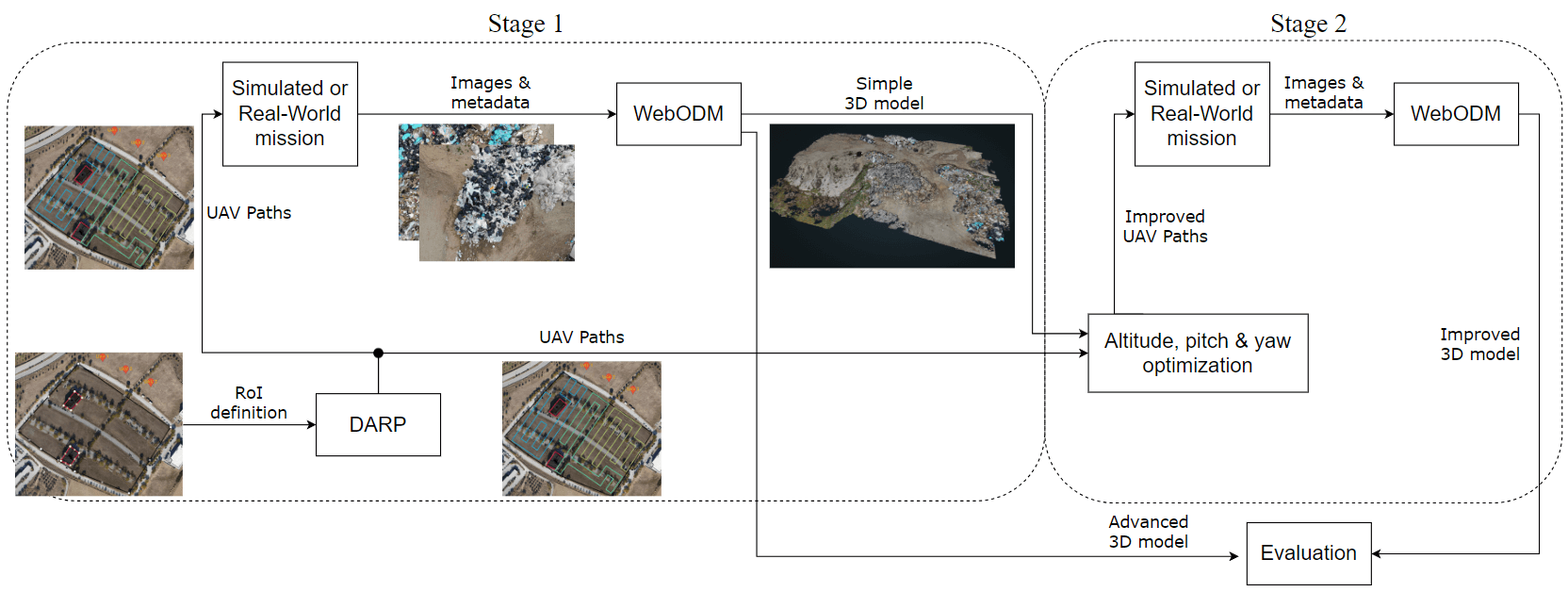}
    \caption{Pipeline of our 3D path planning extension algorithm applied to DARP. Stage 1 produces 2D paths from the baseline algorithm and an initial 3D model of the RoI. These are fed into Stage 2 to produce improved, refined and terrain-aware waypoints in the 3D space. Images used in this figure are exemplary. }
    \label{fig:methodology_diagram}
\end{figure*}

We propose an offline 3D path planning method (see Fig. \ref{fig:methodology_diagram}) that leverages a pre-scanned 3D model to generate optimized UAV viewpoints. The process unfolds in two phases. 

First, a standard path-planning algorithm, in our case DARP, generates 2D flight paths for each UAV. These paths are executed either in simulation or in the field while the UAVs capture images at regular intervals to achieve a desired image overlap. These images then are processed by photogrammetry software, which generates an initial 3D model of the region, hence the dependency of a pre-scanned area. This initial scan is conducted using simpler, more dispersed paths, requiring fewer images and significantly less time. This reduced image count is intentional because the goal is to generate a rough approximation of the environment, not a detailed final model\footnote{This initial 3D model should not be confused with the comparison model, which is produced using more images for evaluation purposes}.

In the second stage, the 3D model and 2D paths that were generated are inputted into a component that adjusts the altitude of the waypoints and then calculates optimal camera angles for each of them in order to improve 3D reconstruction. A mission is then carried out using these optimized paths and camera angles, capturing significantly more images than the first scan. Finally, to produce the improved 3D model, captured images are given as input to the same photogrammetry software.

We refine viewpoints with two distinct processes: altitude adjustment and yaw,pitch angle adjustment. 

\subsection{Path Adjustment}

\begin{figure}[b!]
    \centering
    \includegraphics[width=0.8\linewidth]{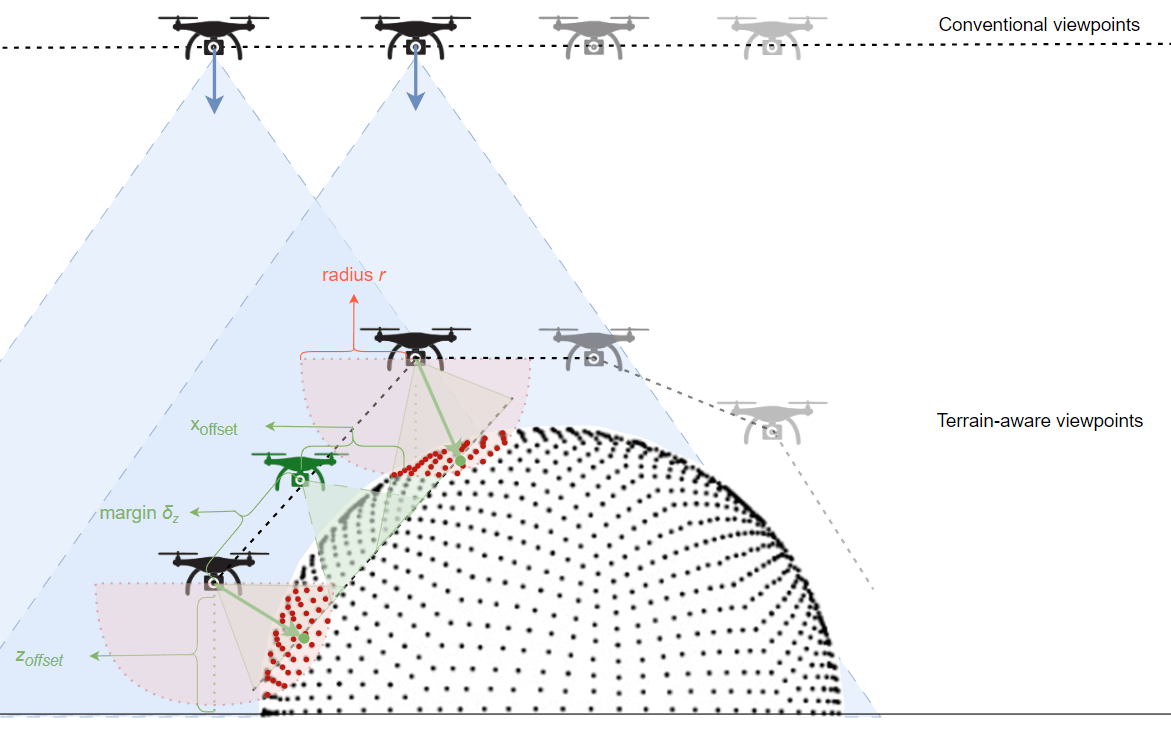}
    \caption{Illustration of viewpoint adaptation to terrain morphology. The drone’s altitude is adjusted to maintain a fixed vertical offset \( z_{\mathrm{offset}} \) from the ground beneath the waypoints. An $x_{offset}$ is also defined to preserve a distance from structures in the same horizontal plane as the UAV. Camera pitch and yaw are then set according to points falling within a downward-facing hemisphere of radius \( r \). Complementary waypoints (green drone) are introduced to ensure better image overlap by a margin \( \delta_z \).}
    \label{fig:DARP3D_paths}
\end{figure}

The algorithm alters the DARP paths vertically, so that they follow the contour of the terrain. This process begins by modifying the altitude of each waypoint, using data from the point cloud. For each waypoint, an iterative search is performed via a K-dimensional tree structure that searches a point in the point cloud which matches the waypoint's \( x \) and \( y \) coordinates (see Alg.\ref{alg:position}). Usually, a point with the exact matching coordinates does not exist in the point cloud, therefore a tolerance is applied in the search. With each iteration, the tolerance increases until at least one point is found. Typically, the higher the density of the point cloud, the lower the tolerance. Once the terrain elevation is determined, a user-defined vertical offset is added to the average z-coordinate of the found points to ensure the drone maintains a safe distance from the ground. Thus, the paths take the shape of the terrain.

\begin{algorithm}
    \caption{Adjust the altitude of every waypoint}
    \begin{algorithmic}[1] 
        \REQUIRE An array $P = \{P_{1}, P_2, P_3, ..., P_n \}$ where $n$ is the number of drones and $p \in P_i$ is each waypoint in the WGS84 system.
        
        An array $M$ representing the 3D model as a point cloud.
        $M_{tol}$ represents the found points.
        
        \STATE Define $tol$, $z_{offset}$
        \FORALL{$P_i \in P$}
            \FORALL{$p \in P_i$}
                \STATE Initialize $M_{tol} = \emptyset $
                \STATE Transform point $p$ to local coordinates $p^l$
                \WHILE{$M_{tol} = \emptyset $}
                    \STATE $M_{tol} \leftarrow KDtree(M, p_x^l, p_y^l, tol)$
                    \STATE $tol \leftarrow tol + \Delta tol$
                \ENDWHILE
                \STATE $p_{avg}^l \leftarrow avg(M_{tol})$
                \STATE $p_z \leftarrow WGS84(p_{avg_z}^l + z_{offset})$
            \ENDFOR
        \ENDFOR
    \end{algorithmic}
    \label{alg:position}
\end{algorithm}

Since DARP does not inherently account for height variations, large gaps can form between waypoints in areas with significant elevation changes. To create a well adjusted path, the algorithm adds new waypoints by traversing the path using a set stride called $step$. 
While traversing the path, at each step, if the height difference between the current point and the waypoint exceeds a certain preset margin $\delta_z$, then a new waypoint is added. 

\subsection{Camera Angles Optimization}

After completing the path adjustment, the algorithm proceeds to calculate optimal camera angles (Alg.\ref{alg:compute_angles}) .

For each waypoint, the algorithm constructs a downward-facing hemisphere centered at the waypoint’s coordinates, starting with a small initial radius $r$ (see Fig.\ref{fig:DARP3D_paths}). This hemisphere is restricted to points at or below the drone’s altitude, as the camera —mounted below the drone— cannot point upward without risking obstruction by the drone itself. The algorithm then searches this hemisphere for points within the pre-scanned 3D point cloud of the environment.

If points are detected within the initial hemisphere, the algorithm calculates their average altitude, terminates the search and selects the point closest to the average altitude of the detected cluster. If no points are found, the radius of the hemisphere is incrementally increased by $\Delta r$, and the search repeats until at least one point is detected. The direction from the waypoint to this selected point defines a favorable camera orientation for that location. Finally, the appropriate yaw and pitch angles are calculated to align the camera with the target.

\begin{algorithm}
\small
    \caption{Compute camera angles for each waypoint}
    \begin{algorithmic}[1]
        \REQUIRE An array $M$ representing the 3D model as a point cloud.
        
        $M_{sph}$ and $M_{h}$ represent the points found inside the sphere and hemisphere respectively.
        \STATE Initialize list of camera angles $A$
        \FORALL{\( P_i \in P \)}
            \FORALL{\( p_j \in P_i \)}
                \STATE Initialize search radius \( r = r_0 \)
                \STATE Initialize hemisphere empty array $M_h$
                \WHILE{\( M_{h} = \emptyset \)}
                    \STATE \( M_{sph} \leftarrow SphereSearch (M,p_j,r) \) 
                    \STATE \( M_{h} \leftarrow \{p_h \in M_{sph} | p_{h_z} < p_{j_z}\}\) (Keep only points below x,y plane, i.e. inside the bottom hemisphere)
                    \STATE Expand search radius: \( r \leftarrow r + \Delta r \)
                \ENDWHILE
                \STATE $p_{avg} \leftarrow avg(M_h)$
                \STATE Find point \( p_{\text{closest}} \) where:
                \[
                p_{\text{closest}} = \arg \min_{p_h \in M_h} |p_{h_z} - p_{avg_z}|
                \]
                \STATE Compute yaw angle \( \psi_d \)
                \STATE Compute pitch angle \( \theta_c \)
                \STATE Store angles \( (\psi_d, \theta_c) \) in $A$
            \ENDFOR
        \ENDFOR
        \RETURN $A$
    \end{algorithmic}
    \label{alg:compute_angles}
\end{algorithm}

Computational overhead is of $\mathcal{O}(n)$ linear time complexity, where $n$ represents the number of waypoints in the longest path. In our tests - including dense path configurations within a \SI{5850}{\metre\squared} RoI - full computation consistently completed within 2 minutes.

\section{Results}
\label{sec:results}

\subsection{Simulation Setup}
The tools used during development and evaluation of the algorithm include:

\begin{itemize}
    \item OpenDroneMap(ODM)\cite{ODM}: The open-source photogrammetry tool of choice. Its open API, code and Docker version allow easy adoption in the workflow.

    \item AirSim\cite{AirSim}: A drone simulator that functions within the Unreal Game Engine. The provided python API and RealWorld2AirSim-DARP\cite{Aliki}, which is an existing port of the DARP algorithm for AirSim, are used for drone control.  

    \item CloudCompare\cite{CC}: An open-source tool for 3D data comparison. In this work, it is used to compare the original 3D model with those constructed utilizing DARP and DARP-3D mission images via cloud-to-cloud analysis.  
\end{itemize}  

\begin{figure}[h!]
    \centering
    \begin{subfigure}[b]{0.49\linewidth} 
        \centering
        \includegraphics[width=0.9\columnwidth, height=0.55\columnwidth]{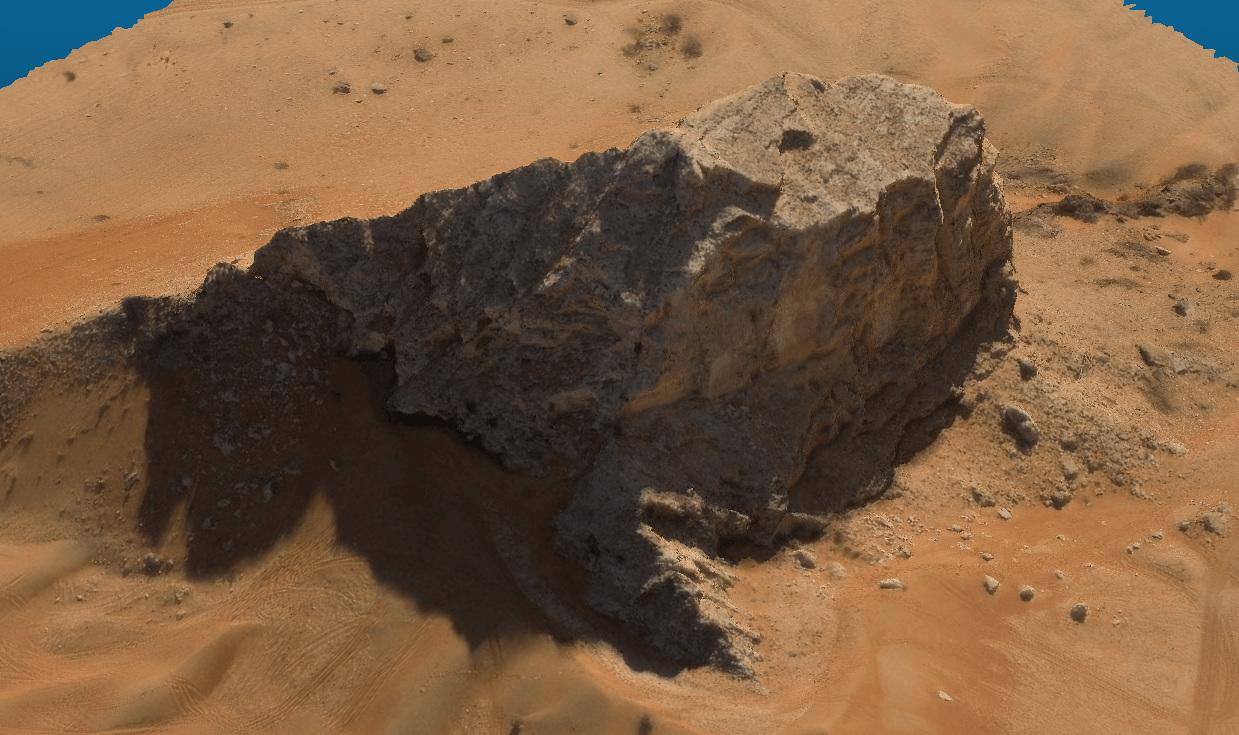} 
        \caption{Dubai Rock}
        \label{fig:model_Dubai}
    \end{subfigure}
    \hfill
    \begin{subfigure}[b]{0.49\linewidth} 
        \centering
        \includegraphics[width=0.9\columnwidth, height=0.55\columnwidth]{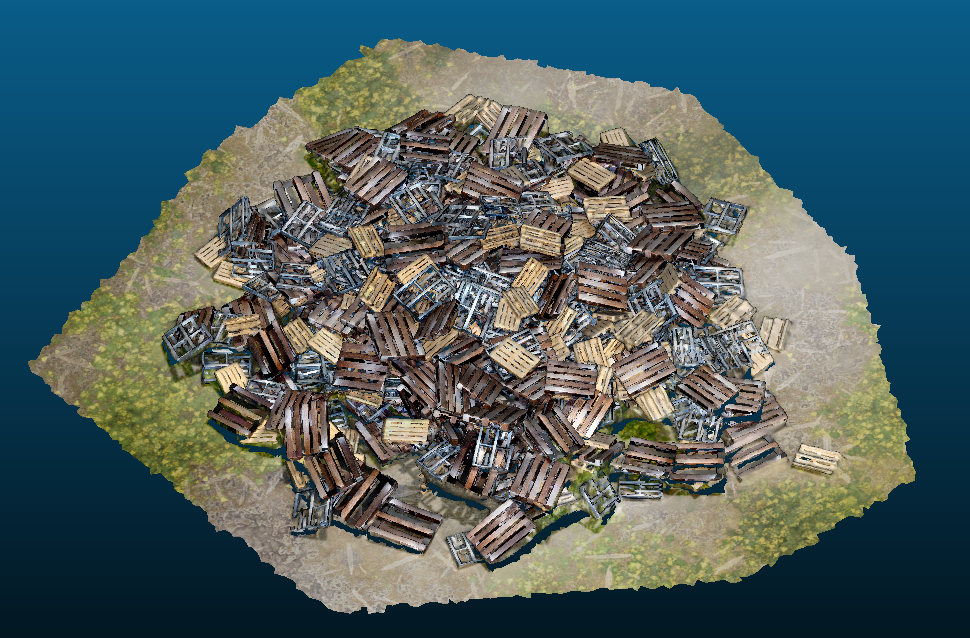} 
        \caption{Pallet Pile}
        \label{fig:model_Pile}
    \end{subfigure}
    \vskip\baselineskip
    \begin{subfigure}[b]{0.49\linewidth} 
        \centering
        \includegraphics[width=0.9\columnwidth, height=0.55\columnwidth]{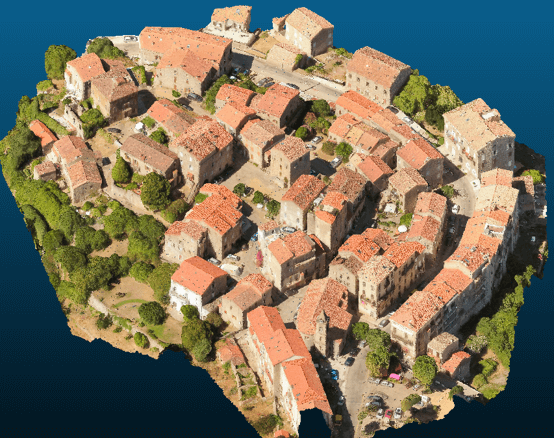} 
        \caption{Village}
        \label{fig:model_Village}
    \end{subfigure}
    \hfill
    \begin{subfigure}[b]{0.49\linewidth} 
        \centering
        \includegraphics[width=0.9\columnwidth, height=0.55\columnwidth]{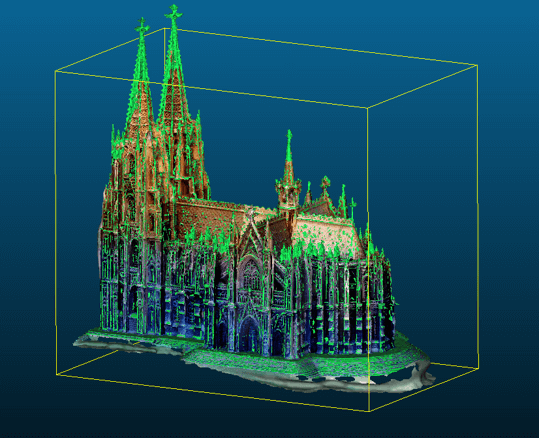} 
        \caption{Cathedral}
        \label{fig:model_Cathedral}
    \end{subfigure}
    \caption{The four 3D models of chosen simulated testbeds. With the exception of (b) which was synthetically generated, (a), (c) and (d) are publicly available and were constructed via photogrammetry. In Fig.\ref{fig:real_comparison} we show the 5th testbed which was captured from the real-world.}
    \label{fig:synthetic_testbeds}
\end{figure}

To execute the missions, four publicly available 3D models (Fig.\ref{fig:synthetic_testbeds}) are used as environments in Unreal Engine, each corresponding to a real-world example: Dubai Rock\cite{ROCK}, Pallet Pile \cite{LFG}, Village \cite{VILLAGE}, Cathedral \cite{CATHEDRAL}

\begin{table}
\caption{Metrics on all testbeds}
\label{table:results}
\centering
\resizebox{\columnwidth}{!}{%
\begin{tabular}{|c|c|c|c|c|c|c|}
\hline
\multicolumn{7}{|c|}{\textbf{Rock}}\\
\hline
 & \multicolumn{2}{|c|}{\textbf{Precision (\%)}} & \multicolumn{2}{|c|}{\textbf{Recall (\%)}} & \multicolumn{2}{|c|}{\textbf{F1-Score (\%)}}\\
\hline
 & 5cm & 10cm & 5cm & 10cm & 5cm & 10cm\\
\hline
DARP & 69.63 & 92.03 & 67.55 & 90.64 & 68.57 & 91.33\\
\hline
DARP (60\degree) & 61.89 & 90.80 & 56.86 & 82.63 & 59.27 & 86.52\\
\hline
DARP-3D & \textbf{82.65} & \textbf{97.45} & \textbf{81.97} & \textbf{96.66} & \textbf{82.31} & \textbf{97.05}\\
\hline
Improvement & \cellcolor[RGB]{191,245,191} 13.02 & \cellcolor[RGB]{228,250,228} 5.42 & \cellcolor[RGB]{184,244,184} 14.42 & \cellcolor[RGB]{224,250,224} 6.02 & \cellcolor[RGB]{187,244,187} 13.74 & \cellcolor[RGB]{226,250,226} 5.72 \\
\hline
\hline
\multicolumn{7}{|c|}{\textbf{Cathedral}}\\
\hline
DARP & 30.58 & 57.83 & 16.07 & 29.90 & 21.07 & 39.42\\
\hline
DARP (60\degree) & 34.22 & 61.62 & 20.11 & 35.93 & 25.33 & 45.39\\
\hline
DARP-3D & \textbf{71.07} & \textbf{85.46} & \textbf{60.44} & \textbf{75.00} & \textbf{65.33} & \textbf{79.89}\\
\hline
Improvement & \cellcolor[RGB]{29,128,29} 40.49 & \cellcolor[RGB]{110,206,110} 27.63 & \cellcolor[RGB]{4,104,4} 44.37 & \cellcolor[RGB]{0,100,0} 45.10 & \cellcolor[RGB]{4,104,4} 44.26 & \cellcolor[RGB]{29,128,29} 40.47 \\
\hline
\hline
\multicolumn{7}{|c|}{\textbf{Village}}\\
\hline
DARP & 7.60 & 30.30 & 6.96 & 23.99 & 7.27 & 26.78\\
\hline
DARP (60\degree) & 11.21 & 48.78 & 10.51 & 42.04 & 10.85 & 45.16\\
\hline
DARP3D & \textbf{23.69} & \textbf{70.43} & \textbf{21.15} & \textbf{55.02} & \textbf{22.35} & \textbf{61.78}\\
\hline
Improvement & \cellcolor[RGB]{175,242,175} 16.09 & \cellcolor[RGB]{31,130,31} 40.13 & \cellcolor[RGB]{184,244,184} 14.19 & \cellcolor[RGB]{88,184,88} 31.03 & \cellcolor[RGB]{181,243,181} 15.08 & \cellcolor[RGB]{63,160,63} 35.00 \\
\hline
\hline
\multicolumn{7}{|c|}{\textbf{Pallet Pile}}\\
\hline
DARP & \textbf{79.93} & \textbf{91.64} & 36.90 & 45.82 & 50.49 & \textbf{61.09}\\
\hline
DARP (60\degree) & 76.75 & 89.54 & 34.42 & 43.24 & 47.53 & 58.32\\
\hline
DARP-3D & 75.38 & 87.80 & \textbf{38.27} & \textbf{46.71} & \textbf{50.77} & 60.98\\
\hline
Improvement & \cellcolor[RGB]{252,230,230} -4.55 & \cellcolor[RGB]{252,234,234} -3.84 & \cellcolor[RGB]{228,250,228} 1.37 & \cellcolor[RGB]{228,250,228} 0.89 & \cellcolor[RGB]{238,250,238} 0.28 & \cellcolor[RGB]{250,240,240} -0.11 \\
\hline
\end{tabular}
}
\end{table}

\subsection{Evaluation}
\label{subsec:evaluation}

\begin{figure*}[t!]
    \centering
    \begin{subfigure}[b]{0.22\textheight}
        \centering
        \includegraphics[width=\columnwidth]{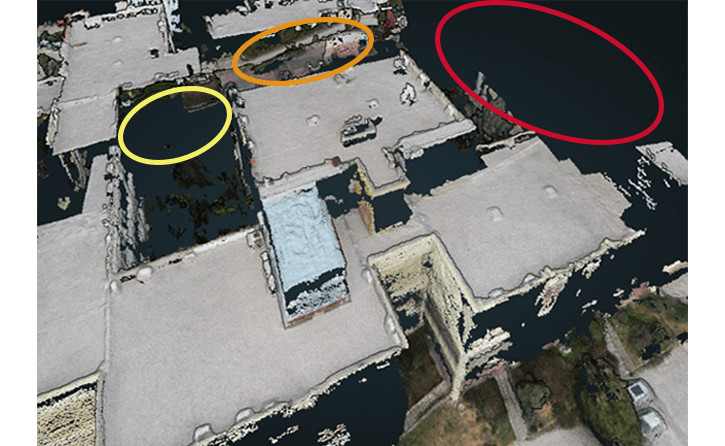}
        \caption{Baseline algorithm 3D model}
        \label{fig:DARP_image}
    \end{subfigure}
    \hfill
    \begin{subfigure}[b]{0.22\textheight}
        \centering
        \includegraphics[width=\columnwidth]{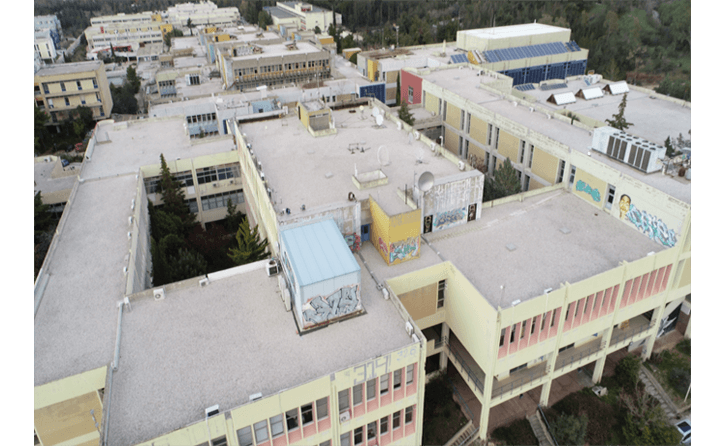}
        \caption{Real world testbed}
        \label{fig:real_image}
    \end{subfigure}
    \hfill
    \begin{subfigure}[b]{0.22\textheight}
        \centering
        \includegraphics[width=\columnwidth]{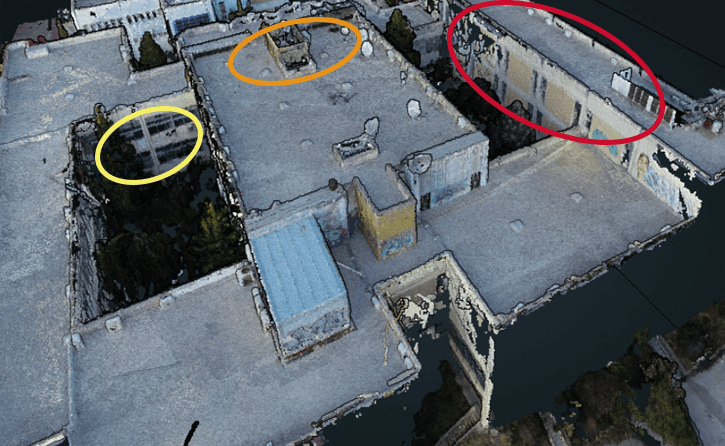}
        \caption{Extended approach 3D model}
        \label{fig:DARP3D_image}
    \end{subfigure}
    
    \caption{Qualitative evaluation of real-life experiment (b). Baseline algorithm (a), and extended approach (c) 3D models were produced. The largest gaps of points are highlighted with colored circled. Many more gaps are visibly filled in this comparison.}
    \label{fig:real_comparison}
\end{figure*}

For each of the aforementioned synthetic models, a simulated flight was performed with AirSim after inputting each original model inside an Unreal Engine environment. After capturing images, ODM produces a digital twin of the model. ODM parameters are kept constant across each testbed.

While differences between the reconstructed models can be seen with a naked eye (see Fig.\ref{fig:viewpoints}), CloudCompare will be used to provide measurable evaluation metrics. To ensure a fair evaluation, both methods will utilize the same number of total images. For instance, if 250 images are used for the DARP algorithm, then its extension, DARP-3D, will use 50 images in the first mission and 200 in the second, totaling another 250 images. The splitting is arbitrary. 
The resulting models are then compared against the original model (i.e. the ground truth).

The following steps outline the evaluation process used:

\begin{enumerate}
	\item Input the three meshes (of DARP, DARP-3D and the GroundTruth) in CloudCompare.
	\item Closely align the meshes in rotation, location, and scale by manually selecting corresponding points, followed by refinement with the Iterative Closest Point (ICP) algorithm.
	\item Sample 5 million points from each mesh.
	\item Calculate the cloud-to-cloud\cite{C2C} distance between the reconstructed 3D models and the ground truth.
	\item Compare the resulting metrics at equivalent thresholds of $5cm$ and $10cm$.
\end{enumerate}

We maintain fairness between comparisons by using the same parameters for ICP alignment of step 2 across all testbeds. A $5cm$ threshold means that we are interested in the percentage of points that have a maximum distance of 5 centimeters from the ground truth, for precision, or vice versa, for recall (see Table \ref{table:results}). This also means that the higher the value the more lenient the measurement becomes. We chose $5cm$ and $10cm$ arbitrary since we believe it is a reasonable objective. Most 2D path planners allow to change the camera angle to a fixed angle which theoretically improves reconstruction, and for this reason, we added DARP with 60° for the comparison. In Table \ref{table:results} the improvement of DARP-3D compared to DARP is highlighted with a red (for negative values) to green (for positive values) gradient. 

In almost all testbeds, our approach greatly improved the baseline. The exception of the Pallet Pile testbed shows that even in the case of a relatively level terrain, our algorithm performs well and similar to baseline. The cause is mainly the size of the pile. Smaller piles consistently tend to cause increased volumetric error \cite{StockpileReview} which is a direct consequence of error in the 3D reconstruction. Furthermore, smaller piles are more sensitive to parameters definition. Perhaps increasing path density and lowering the elevation offset could improve reconstruction accuracy, but this would likely lead to unrealistic computational demands for such a small area. 

\subsection{Real-World Showcase} 
To validate that our method can be easily adopted in real scenarios and improve existing path planners, we performed a qualitative test flight. The flights were conducted in the School of Mining \& Metallurgical Engineering, National Technical University of Athens, Greece (Fig.\ref{fig:real_image}).
All flights were executed in two days for DARP and DARP-3D respectively. For each, roughly 140 images were captured, since a large disparity in the number of images would give an unfair advantage to one approach over the other. Three DJI Phantom 4 Pro UAVs were employed for the flights. 

On the first day, DARP was executed at a $80m$ altitude, $3m/s$ speed, $80\%$ sidelap, $80\%$ frontlap, -90° gimbal pitch (by default), and $33\%$ area coverage for each drone (equal distribution) wielding 132 images. Paths can be planned through its online service \cite{DARP_app}, where a RoI was defined as a polygon encompassing the campus. Waypoints were outputted in json files and lastly an Android app that functions through DJI API, was used to upload the json missions on the real UAVs. Timed interval is automatically calculated by pre-defined parameters. All 132 images were then fed into ODM to produce the 3D model which was used only for a visual comparison (Fig. \ref{fig:DARP_image}).

On the second day, DARP-3D was executed in two stages as Fig.\ref{fig:methodology_diagram} illustrates. The first stage involved running DARP without any modifications. Parameters were the same as in day 1 with the exception of frontlap changed at $50\%$. Lower frontlap sacrifices reconstruction quality for reduced image capture. The json mission was uploaded and wielded 31 images. From them, ODM produced a 3D model serving as an abstract representation of the real geometry. In stage 2, both the simple 3D model and 2D paths were leveraged to output advanced DARP-3D viewpoints and frontlap was reverted to $80\%$. The second json mission was uploaded and wielded 146 images. ODM produced the final and improved 3D model seen in Fig.\ref{fig:DARP3D_image}. It is apparent that many previously empty spots, highlighted with colored circles, were filled with our approach. 

\section{Conclusions}
\label{sec:conclusion}
In summary, we presented a simple yet effective algorithm aimed at upgrading two dimensional path planners for enhanced 3D reconstruction. We demonstrated it by extending the popular DARP path planner. The algorithm showed robust performance in environments with significant elevation changes, such as the Rock, Cathedral, and Village testbeds, where vertical features guide effective camera adjustments and 3D path planning. However, in low-elevation scenarios like the Pile testbed —characterized by small vertical variation— the benefits are limited, with results comparable to the baseline. 

For future work, we aim to better address the challenge of inconsistent image overlap. Gaps in the reconstructed model can persist even when regions are nominally covered, often due to abrupt camera angle differences between waypoints that lead to insufficient overlap. To resolve this, we could capture images at more regular intervals while simultaneously adjusting the yaw and pitch at a smooth and constant rate from waypoint to waypoint. Gradual angle change can guarantee high overlap but is technically challenging due to JSON file format restrictions.

\section*{ACKNOWLEDGMENT}
\label{sec:acknowledgement}

This research has received funding by the European Commission through Horizon Europe project PERIVALLON under Grant Agreement ID No 101073952.

\sloppy
\balance{

}

\end{document}